\documentclass[letterpaper,10pt,conference]{ieeeconf}
\IEEEoverridecommandlockouts
\usepackage{cite}
\usepackage{microtype}
\usepackage[hidelinks]{hyperref}

\usepackage{amsmath,amssymb}
\usepackage{graphicx}
\usepackage{booktabs}
\usepackage{multirow}
\usepackage{placeins}
\usepackage{flafter}
\usepackage{etoolbox}

\patchcmd{\thebibliography}{\footnotesize}{\fontsize{8pt}{8.2pt}\selectfont}{}{}

\AtBeginEnvironment{thebibliography}{%
    \setlength{\itemsep}{0pt}%
    \setlength{\parskip}{0pt}%
    \setlength{\parsep}{0pt}%
}

\title{\LARGE \bf
CoRelNav: Collaborative Relational Navigation for Multi-Robot Spatially Constrained Semantic Navigation}

\author{Jinyu He$^{\#}$, Zihao Mao$^{\#}$, Haonan Jin, Mengyin Fu, and Wenjie Song$^{*}$%
\thanks{This work was partly supported by the National Natural Science Foundation of China (Grant No.~62373052) and the Beijing Natural Science Foundation (Grant No.~4252051).}%
\thanks{The authors are with the School of Automation, Beijing Institute of Technology, Beijing 100081, China.}%
\thanks{$^{\#}$~These authors contributed equally to this work. $^{*}$~Corresponding author: Wenjie Song (email: \texttt{songwj@bit.edu.cn}).}}

\hypersetup{
  pdftitle={CoRelNav: Collaborative Relational Navigation for Multi-Robot Spatially Constrained Semantic Navigation},
  pdfauthor={Jinyu He, Zihao Mao, Haonan Jin, Mengyin Fu, Wenjie Song}
}

\begin{document}

\maketitle
\begin{abstract}
Spatially constrained semantic navigation requires robots to identify targets specified not only by semantic categories but also by relations to surrounding objects. In unknown environments, resolving such goals requires efficient exploration together with sufficient target and contextual evidence for reliable relation verification. Existing methods leave relation-aware verification and multi-robot collaboration largely disconnected: relational navigation is predominantly single-agent, while multi-robot systems seldom coordinate distributed observations for instance-specific relation verification. We propose CoRelNav, whose core is coupling task-conditioned multi-robot exploration with candidate-driven collaborative verification. A spatial-semantic field converts task constraints, scene nodes, and object features into exploration utility; as candidate information accumulates, robots are reallocated toward complementary evidence under team navigation costs, while instance-consistent observations are aggregated across topology nodes. This coupling reduces redundant search and enables relation hypotheses to be resolved from distributed partial evidence that independent exploration or isolated-view verification can leave ambiguous. Experiments in photorealistic simulation demonstrate consistent improvements over representative baselines, with ablations validating the proposed exploration and verification mechanisms. We further deploy the complete system on two physical mobile robots, demonstrating its applicability to real-world collaborative navigation.
\end{abstract}

\begin{figure}[!t]
    \centering
    \includegraphics[width=\columnwidth]
    {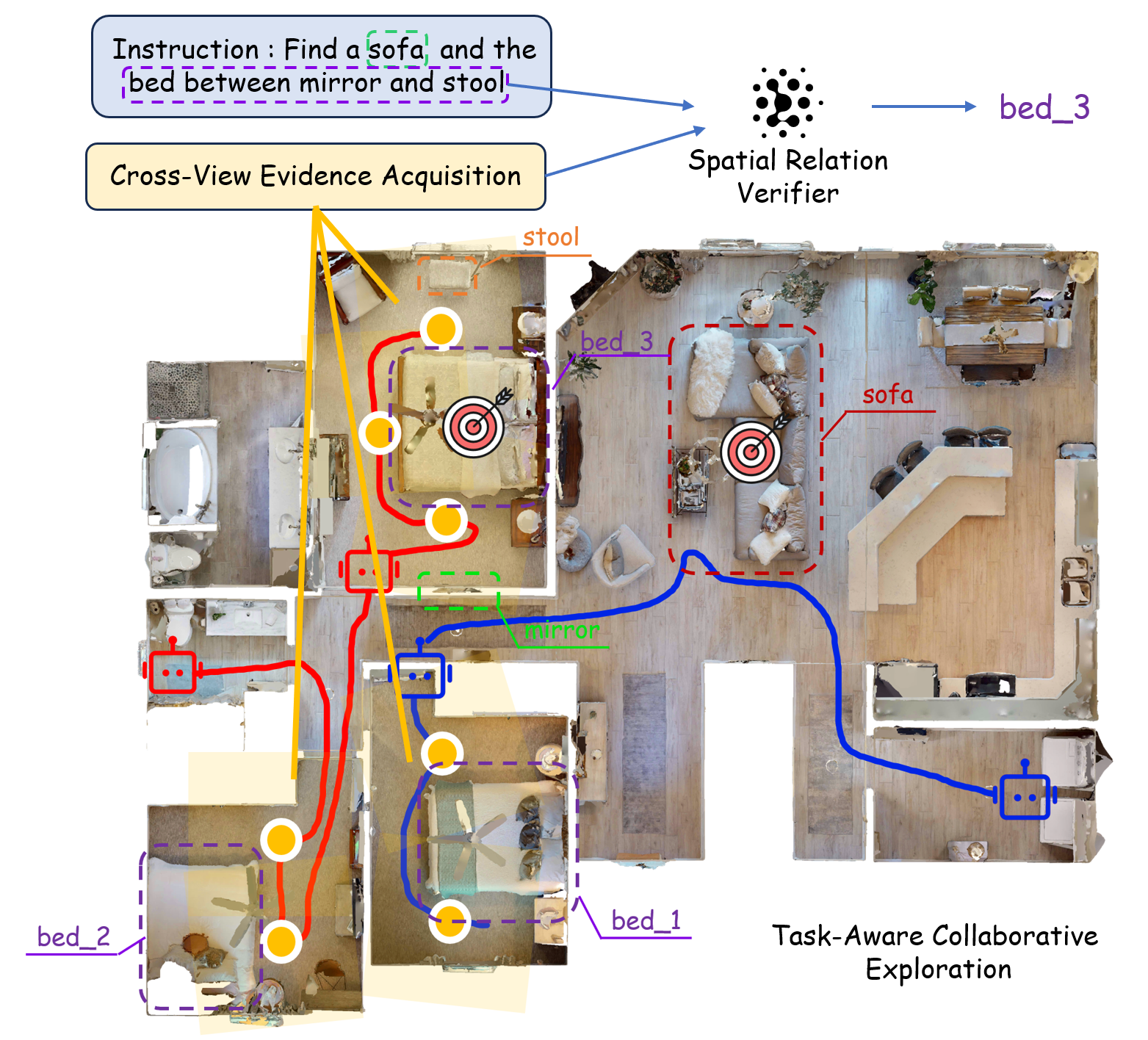}
    \caption{Example of CoRelNav. Given concurrent
    category-level and relational goals, two robots collaboratively
    explore task-relevant regions and acquire cross-view evidence
    around candidate instances. The spatial relation verifier
    aggregates complementary observations around the same candidate
    and confirms \texttt{bed\_3} as the bed between the mirror and the
    stool, while the sofa is handled as a category-level target.}
    \label{fig:intro_overview}
\end{figure}

\section{Introduction}
\label{sec:introduction}

In real-world service, inspection, and assistance scenarios, desired objects are often specified not only by category but also by spatial relations to surrounding objects, especially when multiple same-category instances exist. When several goals must be handled in an unfamiliar environment, a robot team can search different regions in parallel and contribute complementary observations, making collaborative spatially constrained navigation important for efficient task completion and natural human–robot interaction.

Semantic navigation requires embodied robots to locate task-relevant targets in previously unseen environments. Beyond conventional object-goal navigation~\cite{chaplot2020semexp}, practical language instructions may specify a target through spatial relations to surrounding objects, such as \emph{a bed between a mirror and a stool}. In such spatially constrained semantic navigation tasks, detecting a semantically plausible instance is insufficient: the robot must also discover the relevant contextual objects and acquire enough spatial evidence to determine whether the complete relational description is satisfied. Since the target, its supporting objects, and useful contextual cues may be distributed across the environment, multiple robots can search different regions in parallel and share complementary observations, allowing evidence collected by one robot to support another robot's candidate verification. Such collaboration can reduce redundant search and accelerate relation-aware target confirmation, especially when several goals are handled concurrently, while also introducing coordination and evidence-association difficulties.

Recent semantic navigation methods have progressed from category-conditioned search toward open-vocabulary and language-guided goals. Methods such as VLFM~\cite{yokoyama2024vlfm} and InstructNav~\cite{long2025instructnav} use vision-language priors to guide exploration, while relational navigation approaches including VLN-Game~\cite{yu2026vlngame}, DIV-Nav~\cite{ortega2025divnav}, and Context-Nav~\cite{jang2026contextnav} further distinguish candidate discovery from spatial-relation verification. In parallel, multi-robot exploration has evolved from geometry-driven coordination~\cite{yamauchi1997frontier,burgard2005coordinated} toward semantic collaboration, with methods such as Co-NavGPT~\cite{yu2025conavgpt} using shared maps and vision-language reasoning to assign complementary frontiers. However, these two lines remain largely separate: relational navigation is predominantly single-agent, whereas multi-robot semantic navigation mainly optimizes category-level target discovery and rarely considers multiple spatially constrained goals being resolved concurrently. Consequently, spatially constrained navigation with multiple robots remains insufficiently studied.

This setting presents two distinct challenges. First, the search objective is more complex than either geometric coverage or category-level target discovery. A spatially constrained instruction implicitly defines several task-relevant entities, including the target and its contextual objects, and these entities may lie in different regions of the environment. When multiple goals are active, their relevant regions and progress can further differ over time, making redundant exploration or neglect of informative areas more likely. The central difficulty is therefore how to make collaborative exploration remain efficient while being sensitive to the structure and evolving relevance of spatially constrained tasks. Second, even after plausible target candidates are discovered, reliable confirmation remains difficult under partial observation. The target and supporting objects may not be jointly visible, multiple same-category instances may satisfy only part of the description, and evidence gathered from different viewpoints or robots can be incomplete, inconsistent, or associated with different physical instances. The challenge is therefore not simply to detect a likely object, but to determine whether one specific candidate truly satisfies the required spatial relations from distributed and progressively accumulated evidence.

To address these challenges, we propose CoRelNav, a collaborative relational navigation framework that couples task-aware exploration with candidate-driven spatial verification. CoRelNav integrates language task constraints, topological scene nodes, and object features into a task-conditioned spatial-semantic field that guides robots toward task-relevant regions across concurrent goals. As candidate information accumulates, robot allocation adapts according to complementary evidence and team navigation costs, while instance-consistent observations from multiple viewpoints are aggregated for spatial-relation verification. We evaluate CoRelNav in photorealistic indoor environments under concurrent object and spatially constrained language goals and further deploy the complete system on two physical mobile robots. Experiments show stronger gains when task-relevant evidence is spatially dispersed, improved relational-task completion, and reduced language-model usage through candidate-driven verification.

Our main contributions are summarized as follows:

\begin{itemize}
    \item A task-conditioned spatial-semantic field is proposed to address concurrent multi-task and task-relevant multi-robot exploration by fusing task constraints, scene nodes, and object features.
    \item A candidate-driven collaborative allocation and relational verification mechanism is proposed to address target confirmation under partial observations using multi-view evidence and team costs.
\end{itemize}

\section{Related Work}
\label{sec:related_work}

\begin{figure*}[!t]
    \centering
    \includegraphics[width=\textwidth]{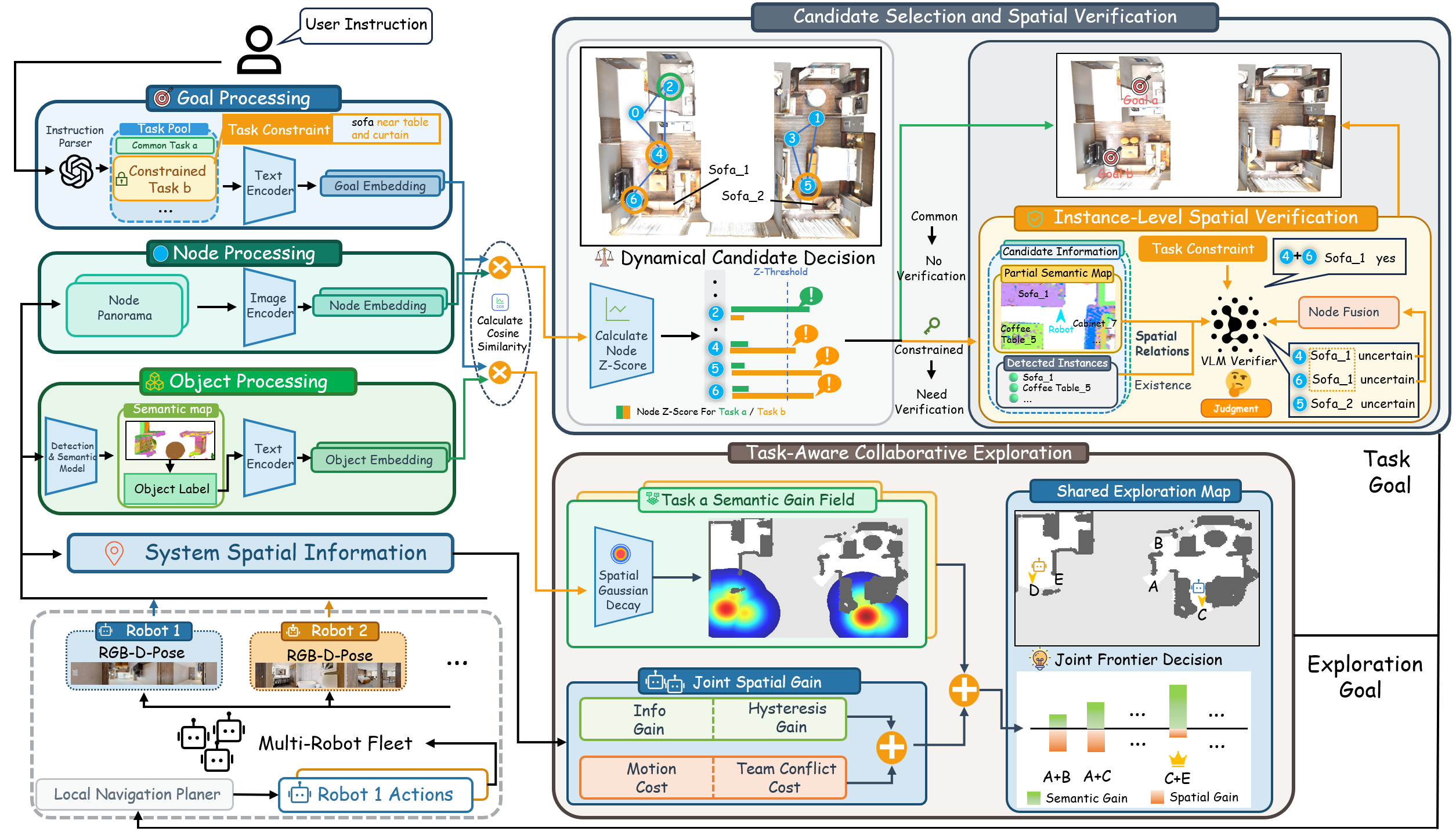}
    \caption{Overview of CoRelNav. Relational instructions, topology
    nodes, and object records form an object-centered task field for
    multi-robot exploration. Acquisition-aware frontier shortlists are jointly
    assigned using team unknown-space gain, object and uncertain-target
    relevance, navigation effort, teammate-history reuse, and navigable
    frontier separation. Candidate hypotheses are then resolved by candidate-level
    spatial verification; uncertain cases trigger complementary views and
    cross-node evidence aggregation before final robot--target navigation.
    An ordered-assignment hysteresis margin limits unnecessary goal switching.}
    \label{fig:framework_overview}
\end{figure*}

\subsection{Language-Conditioned Navigation and Spatial Grounding}

Standard ObjectGoal navigation asks an agent to reach an instance of a specified
category. SemExp builds an episodic semantic map and learns category-conditioned
priors for long-term goal selection with deterministic local
planning~\cite{chaplot2020semexp}, whereas PONI learns frontier potential functions
from passive semantic-map data without interaction-based training for the
``where to look'' component~\cite{ramakrishnan2022poni}. Vision--language methods
broaden the goal interface: CoW combines open-vocabulary image--text localization
with classical exploration~\cite{gadre2023cow}, VLFM ranks frontiers online using
a language-grounded value map~\cite{yokoyama2024vlfm}, InstructNav unifies several
instruction-navigation tasks through Dynamic Chain-of-Navigation~\cite{long2025instructnav},
and UniGoal uses a common graph for object-category, instance-image, and
text-description goals~\cite{yin2025unigoal}. These advances broaden what can be
searched for, but semantic relevance alone does not establish that an instance
satisfies a relational description.

Relational grounding further depends on how spatial context and visual evidence
are represented. VLMaps fuses visual--language features with metric 3D
reconstruction for open-vocabulary and spatial queries~\cite{huang2023vlmaps}.
ConceptGraphs incrementally associates multi-view RGB-D observations into an
open-vocabulary object-centric 3D scene graph~\cite{gu2024conceptgraphs}, while
OVSG grounds free-form, context-aware queries to object instances, agents, and
regions~\cite{chang2023ovsg}. 3D-Mem stores retrievable multi-view Memory and
Frontier Snapshots in an incrementally constructed scene memory~\cite{yang2025threedmem}.
These works contribute scene representations, grounding, or memory mechanisms,
but do not formulate online allocation of incomplete, candidate-specific
relation evidence across a robot team.

Recent navigation systems address relational goals more directly. VLN-Game
builds an object-centric 3D map, identifies promising exploration areas, and uses
game-theoretic vision--language reasoning to select the best-matching
candidate~\cite{yu2026vlngame}. DIV-Nav decomposes a spatial instruction into
object-level queries, intersects their semantic belief maps, and validates the
discovered objects against the original instruction with an LVLM~\cite{ortega2025divnav}.
Context-Nav ranks frontiers from the full description and rejects same-category
distractors through viewpoint-aware 3D relation checking~\cite{jang2026contextnav}.
These methods establish task-guided exploration, instance discrimination, and
relation verification, but evaluate a single agent on one goal or composite query.
Concurrent target--support hypotheses, asynchronous task progress, and
cross-robot relation evidence are outside their formulation.

\subsection{Multi-Robot Exploration and Semantic Collaboration}

\begingroup
\setlength{\parskip}{0pt}
Frontier-based exploration repeatedly selects boundaries between known free
space and unexplored regions~\cite{yamauchi1997frontier}. Burgard et al. coordinate
robots by combining target-reaching cost with frontier utility and discounting
areas expected to be observed by assigned teammates~\cite{burgard2005coordinated}.
Learning-based methods strengthen global allocation: NeuralCoMapping formulates
robot--frontier assignment as neural bipartite graph matching~\cite{ye2022neuralcomapping},
while MAANS learns a transformer-based spatial team planner for rapid joint visual
coverage~\cite{yu2022maans}. Their objectives remain coverage and exploration
efficiency, without valuing observations by their usefulness for resolving a
language-defined relation between specific object instances.

\looseness=1
Semantic collaboration extends this objective beyond coverage. Co-NavGPT uses
a shared representation of explored environments and a vision--language global
planner to assign exploration frontiers to robots for target
search~\cite{yu2025conavgpt}. MCoCoNav combines multimodal reasoning scores with
a global semantic map so that robots choose between frontier points and history
nodes~\cite{shen2025mcoconav}. AnyGoal maintains a shared Bayesian relevance map
across sequential multimodal subtasks and allocates frontiers with a spatial-separation penalty and commitment hysteresis~\cite{james2026anygoal}. Together,
these systems support shared semantic representations and foundation-model
priors for multi-robot target search. Their evaluated tasks, however, use a
common target or a sequence of subtasks rather than several relational tasks
that must be progressed concurrently.

\looseness=1
A gap therefore remains between multi-robot exploration and spatial grounding:
teams must allocate concurrent language-conditioned tasks, seek target and
contextual objects, share instance-level observations, and gather evidence until
each relation can be verified. These requirements are rarely unified in an
online multi-robot setting.
\endgroup

\section{Method}
\label{sec:method}

\subsection{Task Definition}
\label{sec:task_definition}

We present CoRelNav (\textbf{Co}llaborative \textbf{Rel}ational
\textbf{Nav}igation) for concurrent relational navigation with a team
$\mathcal{R}=\{r_1,\ldots,r_N\}$ in an initially unknown environment. The
team receives tasks $\mathcal{T}=\{\tau_1,\ldots,\tau_M\}$, with
\begin{equation}
\tau=(c^{\mathrm{tar}},\mathcal{L}_{\tau}),\qquad
\ell_k=(\rho_k,\pi_k,\mathbf{c}^{\mathrm{sup}}_k)\in\mathcal{L}_{\tau},
\label{eq:task_definition}
\end{equation}
where $c^{\mathrm{tar}}$ is the target category and clause $\ell_k$ contains a
stable identifier $\rho_k$, predicate $\pi_k$, and ordered support-category
tuple $\mathbf{c}^{\mathrm{sup}}_k$; $k$ indexes clauses and tuple length
preserves predicate arity.
For example, ``the bed between a shelf and a window, and under a painting''
becomes $\mathrm{between}(\mathrm{bed},(\mathrm{shelving},\mathrm{window}))$
and $\mathrm{under}(\mathrm{bed},\mathrm{painting})$ after canonicalizing
\textit{shelf} to \textit{shelving}.

\subsection{System Overview}
\label{sec:method_overview}

As illustrated in Fig.~\ref{fig:framework_overview}, goal, node, and object
processing fuse the instruction, panoramas, exact-instance evidence, and
geometry into a shared task-conditioned field. At time $t$, robot $r_i$ adds
$o_t^i=(I_t^i,D_t^i,\mathbf p_t^i)$---color, depth, and sensor pose---to this
representation. The field produces acquisition-aware frontier shortlists that
are jointly ranked using task relevance, complementary coverage, and execution
costs.
Candidate nodes then bind to exact targets: uncertainty triggers complementary
views and instance-consistent aggregation, while verification yields approach
goals. Thus partial cues guide discovery, but exact clauses determine completion.

\subsection{Shared Spatial-Semantic Representation}
\label{sec:shared_representation}

\textbf{Shared map and scene memory.}
The occupancy map $M_t^{\mathrm{geo}}$ defines free and unknown cell sets
$\mathcal X_{\mathrm{free}}$ and $\mathcal X_{\mathrm{unk}}$. The online
instance set $\mathcal I_t=\{\iota_j\}$ stores each exact ID, canonical
category, map-frame geometry, and observation history. A sparse topological
graph $\mathcal G_t=(\mathcal V_t,\mathcal E_t)$ complements this fused map:
each node $v_n$ retains its pose, panorama $\Pi_n$, local semantic evidence,
visible instance set $\mathcal I_n$, and connectivity. This separates
navigation geometry from retrievable, viewpoint-specific relation evidence.

\textbf{Task-conditioned spatial-semantic field.}
Let $q_\tau$ be the full instruction and $d_j$ describe instance $\iota_j$
using its category and nearby observed categories. Aligned encoders give
$\mathbf e_\tau=E_{\mathrm{txt}}(q_\tau)$,
$\mathbf e_n=E_{\mathrm{img}}(\Pi_n)$, and
$\mathbf e_j=E_{\mathrm{txt}}(d_j)$. We define
\begin{equation}
\begin{aligned}
a_{n,j}^{\tau}&=\lambda_n\cos(\mathbf e_\tau,\mathbf e_n)
+\lambda_o\cos(\mathbf e_\tau,\mathbf e_j),\\
H^\tau(\mathbf x)&=\mathbf 1(\mathbf x\!\in\!\mathcal X_{\mathrm{free}})
\max_{n,\,j:\iota_j\in\mathcal I_n}[a_{n,j}^{\tau}]_+
\kappa_\sigma(\mathbf x,\mathbf x_j),
\end{aligned}
\label{eq:semantic_field}
\end{equation}
where $\lambda_n,\lambda_o\geq0$ sum to one, $[u]_+=\max(u,0)$,
$\kappa_\sigma(\mathbf x,\mathbf x_j)=
\exp(-\|\mathbf x-\mathbf x_j\|_2^2/2\sigma^2)$, and $\mathbf x_j$ is the
object center. Here $\mathbf x$ is a queried map location, $\mathbf 1(\cdot)$
is the indicator function, $a_{n,j}^{\tau}$ is the task compatibility of
instance $j$ at node $n$, and $\sigma$ is the field bandwidth. Thus,
$H^\tau$ represents the spatial utility of acquiring evidence relevant to task
$\tau$, rather than merely the semantic likelihood of the target category.
Target, support, and contextual observations can therefore guide exploration
before the complete relation becomes directly verifiable, while exact relation
satisfaction remains deferred to the verification stage. The same node
embedding supplies the proposal score
$s_n^\tau=\cos(E_{\mathrm{txt}}(c^{\mathrm{tar}}),\mathbf e_n)$, separating
region-level exploration guidance from node-level hypothesis selection.

\begin{figure}[!t]
    \centering
    \includegraphics[width=\columnwidth]{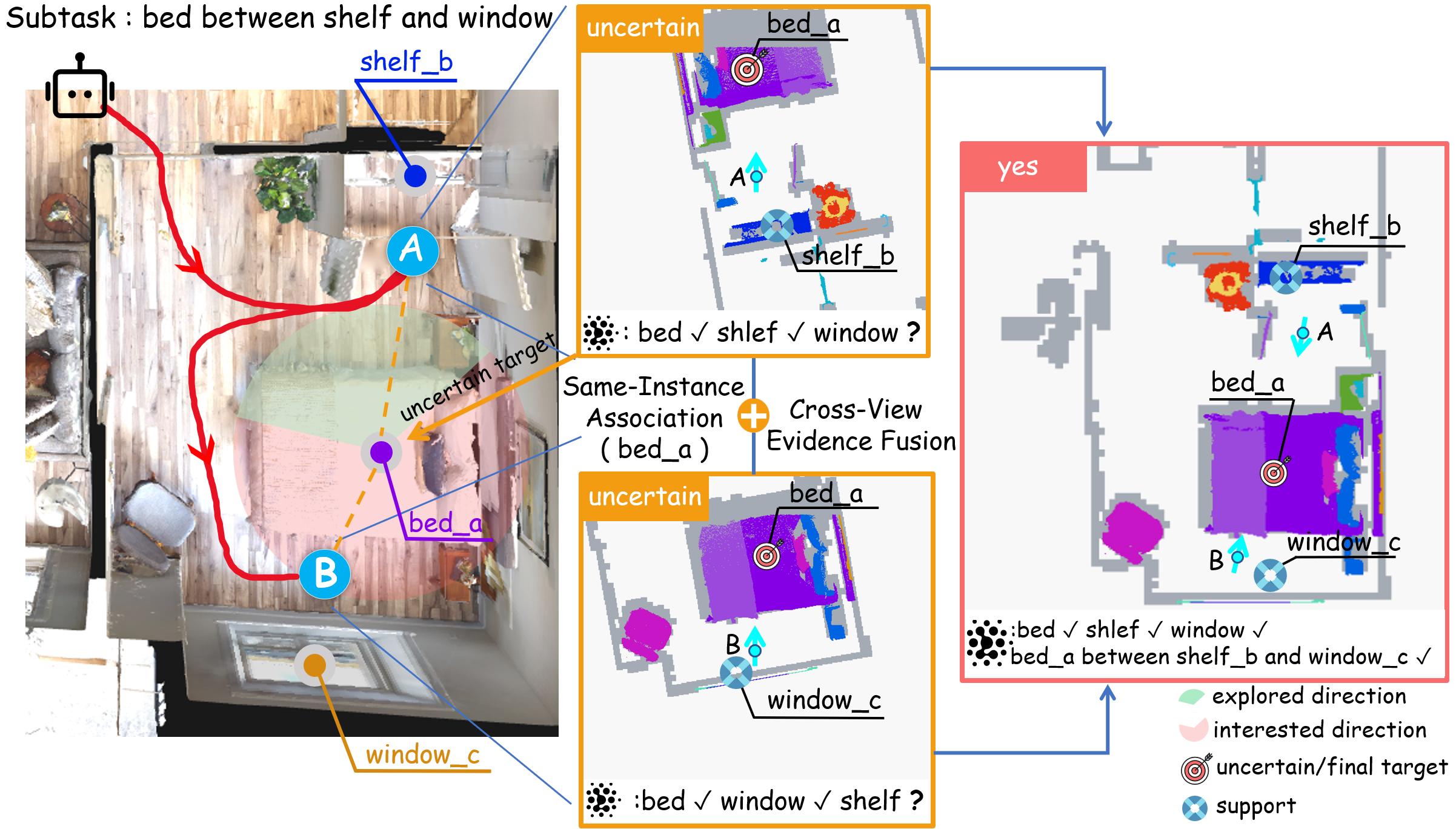}
    \caption{Cross-node complementary evidence aggregation for the subtask
    ``find the bed between a shelf and a window.'' Node A observes candidate
    \texttt{bed\_a} with \texttt{shelf\_b} but lacks \texttt{window\_c}, while
    Node B provides the complementary observation with \texttt{window\_c}.
    Aggregating the two node-level observations around the same candidate
    hypothesis enables the verifier to confirm that \texttt{bed\_a} lies
    between \texttt{shelf\_b} and \texttt{window\_c}.}
    \label{fig:multinode_evidence}
\end{figure}

\subsection{Task-Aware Multi-Robot Collaborative Exploration}
\label{sec:collaborative_exploration}

Let $\mathcal T_t^{\mathrm{act}}$ be the unfinished tasks in the shared pool,
and let $\mathcal F_i^{\mathrm{fea}}$ contain frontiers reachable by robot $r_i$
through known free space with valid clearance and floor consistency. For
$f\in\mathcal F_i^{\mathrm{fea}}$, a bounded collision-truncated rollout
$\widehat\Gamma_i(f)$ predicts visible cells $\operatorname{Vis}(\boldsymbol
\xi)$ from sampled pose $\boldsymbol\xi$. With terminal pose
$\widehat\Gamma_i^{\mathrm{end}}(f)$, define the acquisition gain and
per-robot shortlist score as
\begin{equation}
\begin{aligned}
\mathcal U_i(f)&=\Big(\!\bigcup_{\boldsymbol\xi\in\widehat\Gamma_i(f)}
\operatorname{Vis}(\boldsymbol\xi)\!\Big)\cap\mathcal X_{\mathrm{unk}},\\
G_i(f)&=|\mathcal U_i(f)|+
\eta|\operatorname{Vis}(\widehat\Gamma_i^{\mathrm{end}}(f))
\cap\mathcal X_{\mathrm{unk}}|,\\
E_i^\tau(f)&=w_d^p\bar D_i^\tau(f)+w_o^p\bar O_i^\tau(f)
+w_b^pB_i^\tau(f),\\
P_i(f)&=w_u^p\bar G_i(f)
+\max_{\tau\in\mathcal T_t^{\mathrm{act}}}E_i^\tau(f)\\
&\quad-w_a^p\bar A_i(f)-w_h^p\bar R_i(f),
\end{aligned}
\label{eq:prefilter_utility}
\end{equation}

where $\eta$ weights the terminal-view bonus, $D_i^\tau$ is the directional
semantic prior from the latest topology node, $O_i^\tau$ is the
task-conditioned spatial-semantic field (object heat),
and $B_i^\tau$ rewards a missing view or complementary observation for an
\textsc{Uncertain} candidate. Thus, $E_i^\tau$ is the task-evidence acquisition
utility. $A_i$ is the number of translation and turning primitives on the
feasible A* path, and $R_i$ is the fraction of that path reusing the teammate's
exploration history. A bar denotes normalization by a fixed typical scale, and
all weights are nonnegative. Retaining the top-$K$ candidates yields
$\widehat{\mathcal F}_i$ while preserving nearby, visually complementary
frontiers as distinct options; each retained candidate carries label
$\tau_i(f)=\arg\max_{\tau\in\mathcal T_t^{\mathrm{act}}}E_i^\tau(f)$.
These per-robot scores are used only to construct shortlists; final goals are
selected jointly at the team level.

For active robots $\mathcal A\subseteq\{1,\ldots,N\}$, let
$\Omega_{\mathcal A}\subseteq\prod_{i\in\mathcal A}
\widehat{\mathcal F}_i$ contain only valid, non-duplicate,
navigation-distinct, and sufficiently separated tuples
$\mathbf f_{\mathcal A}=(f_i)_{i\in\mathcal A}$. Below,
$\tau_i=\tau_i(f_i)$ may differ across robots.
\begin{equation}
\begin{aligned}
\mathcal V_i^{\mathrm{end}}(f)
&=\operatorname{Vis}(\widehat\Gamma_i^{\mathrm{end}}(f))
\cap\mathcal X_{\mathrm{unk}},\\
G_{\cup}(\mathbf f_{\mathcal A})
&=\left|\bigcup_{i\in\mathcal A}\mathcal U_i(f_i)\right|
+\eta\left|\bigcup_{i\in\mathcal A}
\mathcal V_i^{\mathrm{end}}(f_i)\right|,\\
\langle X\rangle_{\mathcal A}
&=\tfrac1{|\mathcal A|}\sum_{i\in\mathcal A}X_i(f_i),
\quad X_i\in\{O_i^{\tau_i},B_i^{\tau_i},R_i\},\\
A_{\Sigma}&=\sum_{i\in\mathcal A}A_i(f_i),\\
\langle Z\rangle_{\mathcal A}
&=\frac{\sum_{i<k\in\mathcal A}Z_{ik}}
{\max\{1,\binom{|\mathcal A|}{2}\}},
\quad Z\in\{d_{\mathrm{geo}},C\},\\
Q_{\mathcal A}(\mathbf f_{\mathcal A})
&=w_u^j\bar G_{\cup}+w_o^j\overline{\langle O\rangle}_{\mathcal A}
 +w_b^j\langle B\rangle_{\mathcal A}\\
&\quad+w_s^j\overline{\langle d_{\mathrm{geo}}\rangle}_{\mathcal A}
-w_a^j\bar A_{\Sigma}\\
&\quad-w_h^j\overline{\langle R\rangle}_{\mathcal A}
-w_c^j\langle C\rangle_{\mathcal A},\\
\mathbf f_{\mathcal A}^{*}
&=\arg\max_{\mathbf f_{\mathcal A}\in\Omega_{\mathcal A}}
Q_{\mathcal A}(\mathbf f_{\mathcal A}).
\end{aligned}
\label{eq:joint_assignment}
\end{equation}
Here $\mathcal V_i^{\mathrm{end}}$ is the unknown area visible from the rollout's
terminal pose, $G_{\cup}$ measures non-overlapping team acquisition,
$A_{\Sigma}$ is the summed action cost, and $C$ is pairwise corridor conflict;
angle brackets denote
robot-wise or pairwise means and pairwise terms are zero for a singleton.
Directional semantics is used only for per-robot shortlisting. All unfinished-task
candidates enter this joint solve, allowing different task labels across robots.
The full active set is optimized
whenever feasible; otherwise, all largest feasible subsets are evaluated and
omitted robots retain valid goals. For equal-cardinality assignments, the
previous ordered tuple is retained unless $\Delta Q\geq\Delta_h=2$.

\subsection{Candidate-Centric Spatial Relation Verification}
\label{sec:verification}

\textbf{Candidate decision.}
For each task, node scores are robustly standardized using lower-quantile
baseline $b_\tau=\operatorname{quantile}_{0.30}(\{s_n^\tau\}_n)$, robust deviation scale
$r_\tau=\max(1.4826\operatorname{median}_n|s_n^\tau-b_\tau|,\epsilon)$, and
$z_n^\tau=(s_n^\tau-b_\tau)/r_\tau$. All nodes satisfying $s_n^\tau > \theta_s$ and $z_n^\tau > \theta_z$ are retained in
$\mathcal V_\tau^{\mathrm{cand}}$. Each compatible target
candidate forms a persistent hypothesis
$h=(\tau,\mathrm{id}^{\mathrm{tar}},\mathcal V_h,\mathcal D_h)$ containing its
evidence nodes and observed sectors. Fresh nodes precede retries; each task--node
pair allows at most three attempts with an eight-cycle cooldown after non-transport
failures; transport failures consume no attempt, and positive results are latched.

\textbf{Structured single-node verification.}
For node $v_n$, $E_n(h)$ contains an oriented local occupancy ROI, every
projected object instance, its ID/category index, and vertical cues. With the
original clauses, the structured verifier returns
\begin{equation}
\begin{aligned}
\Psi_{\mathrm{rel}}(E_n(h),\mathcal L_\tau)&\rightarrow
\big(y,\mathrm{id}^{\mathrm{tar}},\\[-1mm]
&\hspace{5mm}\{(\rho_k,\mathbf{id}^{\mathrm{sup}}_k,\nu_k)\}_k\big),
\end{aligned}
\label{eq:verifier}
\end{equation}
where $y\in\{\mathrm{YES},\mathrm{NO},\mathrm{UNCERTAIN}\}$,
$\mathbf{id}^{\mathrm{sup}}_k$ binds ordered support IDs, and $\nu_k$ is the
clause verdict. Acceptance requires the deterministic gate $C(h,E)=1$:
\[
\begin{aligned}
C(h,E)={}&\mathbf 1[y=\mathrm{YES}]\\
&\cdot\prod_{\ell_k\in\mathcal L_\tau}
\Bigl(\mathbf 1[\nu_k=\mathrm{YES}]\\[-1mm]
&\hspace{11mm}\cdot\mathbf 1[\operatorname{Valid}
(\rho_k,\mathbf{id}^{\mathrm{sup}}_k,E)]\Bigr).
\end{aligned}
\]
Here $\operatorname{Valid}$ checks exact IDs and categories, predicate arity,
and required supporting evidence. \textsc{Yes} promotes the target,
\textsc{No} rejects only the node proposal, and \textsc{Uncertain} preserves
the hypothesis and triggers complementary re-observation.

\textbf{Cross-node evidence aggregation.}
Fig.~\ref{fig:multinode_evidence} shows why a single node can be insufficient.
For nodes $\mathcal V_h$ of the same exact-target hypothesis, each node fixes
its pose and oriented $4\,\mathrm m\times4\,\mathrm m$ evidence region
$\operatorname{ROI}_n$. At aggregation time, let $P_j^{\mathrm{live}}(t)$ be
the latest accumulated map-frame point cache of exact instance $j$. We form
\begin{equation}
\begin{aligned}
\mathcal R_h&=\bigcup_{n\in\mathcal V_h}\operatorname{ROI}_n,\\
P_j^{\mathrm{agg}}(t)&=P_j^{\mathrm{live}}(t)\cap\mathcal R_h,\\
E_h^{\mathrm{agg}}&=\operatorname{Render}\!\left(
M_t^{\mathrm{geo}}|_{\mathcal R_h},\{P_j^{\mathrm{agg}}(t)\}_j\right),
\end{aligned}
\label{eq:multi_node_aggregation}
\end{equation}
where $M_t^{\mathrm{geo}}|_{\mathcal R_h}$ is the latest occupancy context
restricted to the union region. Thus evidence observed after an individual
node was created can enter the aggregate, while persistent candidate identity prevents
category-matched but distinct objects from being mixed. Complementary nodes may
supply different clauses while remaining bound to one target hypothesis; the
same structured verifier then evaluates the refreshed aggregate.

\textbf{Final robot--target assignment.}
For each verified instance $\iota_j$, known-free approach poses
$\mathcal A_j$ are sampled outside its observed footprint (and an eligible
supporting carrier when applicable). The cost
$C_{ij}=\min_{a\in\mathcal A_j}\operatorname{cost}_{\mathrm{path}}(\mathbf
x_i,a)$ is the finite A* cost from robot $i$ at map position $\mathbf x_i$;
Hungarian assignment minimizes
total cost over reachable pairs. The task closes when the assigned robot reaches
the verified instance, while remaining robots continue unfinished tasks. This
converts a relation decision into an executable goal and closes the chain.

\section{Experiments}
\label{sec:experiments}

In this section, we evaluate CoRelNav in simulation on object-goal and language-description tasks and deploy it on two mobile robots to assess real-world feasibility.

\begin{figure*}[!t]
    \centering
    \makebox[\textwidth][c]{%
        \begin{minipage}[t]{0.242\textwidth}
            \centering
            \scriptsize\textbf{(a) Step 38}\par\vspace{0.4mm}
            \includegraphics[width=\linewidth,viewport=23 0 897 842,clip]
            {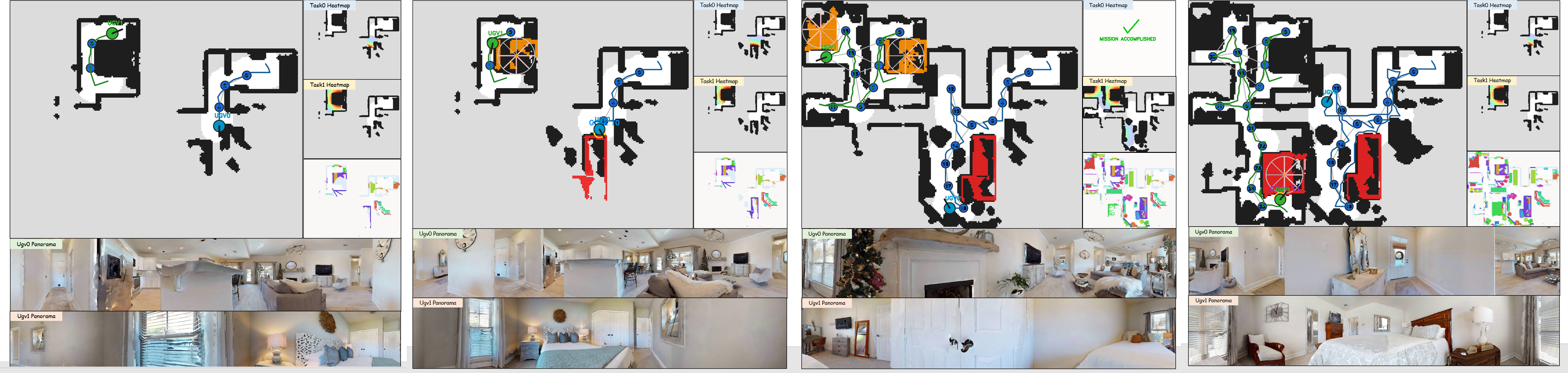}
        \end{minipage}%
        \hfill
        \begin{minipage}[t]{0.242\textwidth}
            \centering
            \scriptsize\textbf{(b) Step 54}\par\vspace{0.4mm}
            \includegraphics[width=\linewidth,viewport=919 0 1793 842,clip]
            {fig4_new_steps_38_54_172_314_v2.png}
        \end{minipage}%
        \hfill
        \begin{minipage}[t]{0.242\textwidth}
            \centering
            \scriptsize\textbf{(c) Step 172}\par\vspace{0.4mm}
            \includegraphics[width=\linewidth,viewport=1794 0 2668 842,clip]
            {fig4_new_steps_38_54_172_314_v2.png}
        \end{minipage}%
        \hfill
        \begin{minipage}[t]{0.242\textwidth}
            \centering
            \scriptsize\textbf{(d) Step 314}\par\vspace{0.4mm}
            \includegraphics[width=\linewidth,viewport=0 16 1261 1231,clip]
            {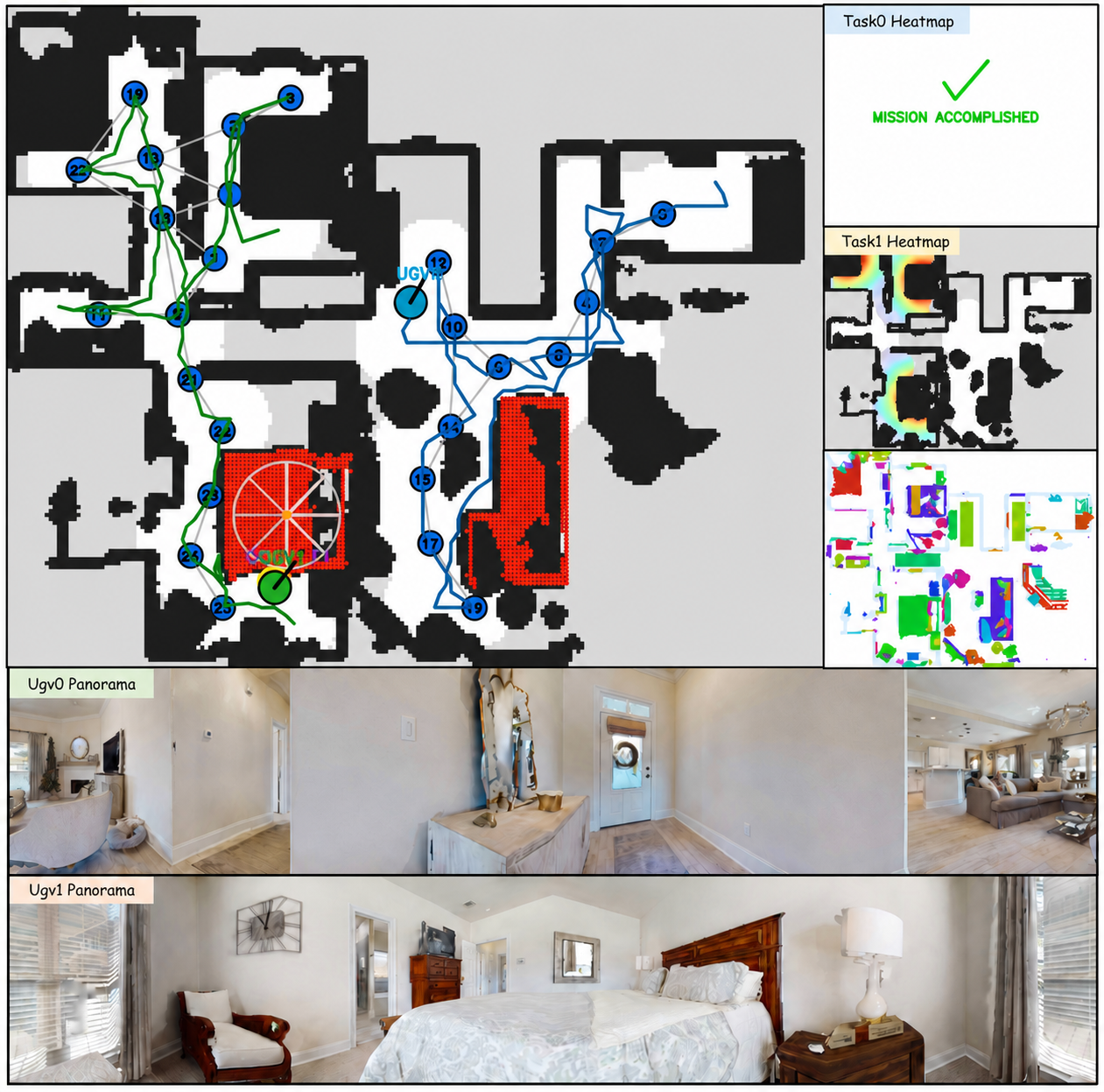}
        \end{minipage}%
    }

    \vspace{-0.5mm}

    \caption{Progression of CoRelNav on the episode ``find a sofa and a bed
    between a stool and a mirror.'' The four panels show (a) verification of
    the first bed at step 38, (b) discovery of the sofa at step 54,
    (c) verification of the second bed at step 172, and (d) verification of
    the third bed and its confirmation as the final target at step 314. In each panel, the main map shows the evolving shared map and robot trajectories, with orange and red point-cloud projections denoting uncertain and final targets, respectively; the side insets show task-conditioned heatmaps, while the bottom rows show the latest panoramic observations from UGV0 and UGV1.}
    \label{fig:simulation_qualitative}
\end{figure*}

\subsection{Simulation Experiment}
\label{sec:simulation_experiment}

\subsubsection{Dataset}
We conduct experiments using high-resolution photorealistic 3D reconstructions
from the HM3D dataset~\cite{ramakrishnan2021hm3d}. Our episode construction
closely follows VLN-Game~\cite{yu2026vlngame}, while adapting it to the
concurrent multi-robot setting. We construct two sets of evaluation episodes.
The first contains unconstrained multi-task episodes, where two object-goal
targets must be located in each scene. The second contains mixed multi-task
episodes combining object goals with language descriptions, with the
instructions primarily emphasizing spatial relations such as \emph{near},
\emph{under}, and \emph{between}. Each episode specifies the scene, two
initial robot poses, two target descriptions, and the corresponding success
viewpoints. The same episode definitions and initial poses are used for all
methods in each comparison. We label an episode same-region when the minimum
geodesic distance between its two target success regions is below $5\,$m,
and cross-region otherwise.

\subsubsection{Settings}
Experiments use Habitat~\cite{savva2019habitat} with $480\times640$ RGB-D
observations and poses, a $0.075\,$m shared occupancy map, instance-aware
topology, and a budget of 600 joint planning steps. Low-level actions are
$0.25\,$m forward and $30^{\circ}$ turns. For external baselines,
the two subtasks are issued sequentially, with the second provided only after
the first terminates; CoRelNav instead maintains both subtasks jointly in a
shared task pool and reallocates team effort as evidence accumulates. To isolate
exploration and reasoning, simulation uses simulator-provided
instance masks and IDs instead of Grounding DINO~\cite{liu2024groundingdino}
and MobileSAM~\cite{zhang2023mobilesam}, while mapping, exploration, evidence
collection, verification, and navigation remain online.
\texttt{qwen3-vl-embedding}~\cite{li2026qwen3vlembedding} provides semantic
guidance and \texttt{qwen3.7-plus}~\cite{alibaba2026qwen37plus} performs
structured relation verification; dual robots share maps, semantic
observations, task states, and topology evidence.

\subsubsection{Evaluation Metrics}
We report SR and an adapted team SPL metric based on~\cite{anderson2018evaluation}
for object, relational, and episode-level completion; an episode succeeds only
when both subtasks and their navigation requirements are completed. SR is
$\frac{1}{N}\sum_{i=1}^{N}S_i$, with $S_i=1$ for a successful episode. Let
$L_i$ be the shortest route from the canonical first-robot start visiting both
target success regions in either order. SPL is computed as
\begin{equation}
\mathrm{SPL}=\frac{1}{N}\sum_{i=1}^{N}
S_i\frac{L_i}{\max(L_i,P_i)},
\label{eq:experiment_spl}
\end{equation}
where $P_i$ is the single-robot path length or the sum of both robot
trajectories for a dual-robot method. We additionally report average
language-model tokens per episode.

\subsubsection{Baselines}
For unconstrained exploration, our Greedy variant of the standard baseline
family evaluated by Visser et al.~\cite{visser2013greedy} selects the nearest
non-conflicting frontier. Cost--Utility~\cite{julia2012costutility} lets each
robot independently maximize normalized unknown-space gain minus weighted
normalized path cost, while Co-NavGPT~\cite{yu2025conavgpt} jointly assigns frontiers using VLM reasoning
over the shared top view. Greedy and Cost--Utility use the same map and
low-level navigation interface as ours, while Co-NavGPT uses the same
observations, episodes, and action budget.

For mixed tasks, Embedding is our concurrent dual-robot system with structured
relation verification disabled; it directly selects the highest-similarity
candidate. VLN-Game (Dual)~\cite{yu2026vlngame} runs two
independent agents in parallel using the original vision-language equilibrium
search and candidate-identification pipeline, without inter-robot information
sharing or coordination.

\subsubsection{Results and Discussion}
\begin{table}[!ht]
\centering
\caption{Performance stratified by spatial distribution.}
\label{tab:spatial_distribution}
{\footnotesize\rmfamily
\setlength{\tabcolsep}{5.0pt}
\renewcommand{\arraystretch}{1.16}
\begin{tabular}{@{}ccccc@{}}
\toprule
\multirow{2}{*}{\textbf{Method}} & \multicolumn{2}{c}{\textbf{Same-region}} & \multicolumn{2}{c}{\textbf{Cross-region}} \\
\cmidrule(lr){2-3}\cmidrule(l){4-5}
& \textbf{SR} $\uparrow$ & \textbf{SPL} $\uparrow$ & \textbf{SR} $\uparrow$ & \textbf{SPL} $\uparrow$ \\
\midrule
Greedy & 0.648 & 0.168 & 0.655 & 0.166 \\
Cost--Util. & 0.675 & 0.169 & 0.672 & 0.175 \\
Co-NavGPT & 0.709 & 0.202 & 0.701 & 0.214 \\
\textbf{CoRelNav} & \textbf{0.716} & \textbf{0.208} & \textbf{0.746} & \textbf{0.252} \\
\bottomrule
\end{tabular}
}
\end{table}

Table~\ref{tab:spatial_distribution} shows that the advantage of CoRelNav
becomes more pronounced when task targets are spatially dispersed.
In same-region episodes, different exploration policies can often reach
overlapping informative areas, leaving limited room for coordination. In cross-region episodes, however, independently attractive frontiers are more likely to cause redundant coverage or concentrate both robots on locally salient regions. CoRelNav’s larger gain therefore supports the intended roles of the task-conditioned spatial-semantic field and joint allocation: the former identifies task-relevant regions, while the latter converts this relevance into complementary assignments under team coverage and navigation cost.

\FloatBarrier

\begin{table}[!ht]
\centering
\caption{Main results in mixed multi-task navigation.}
\label{tab:mixed_results}
{\footnotesize\rmfamily
\setlength{\tabcolsep}{2.6pt}
\renewcommand{\arraystretch}{1.16}
\begin{tabular}{@{}lccccc@{}}
\toprule
\multirow{2}{*}{\textbf{Method}} & \multicolumn{1}{c}{\textbf{Object}} & \multicolumn{1}{c}{\textbf{Relational}} & \multicolumn{3}{c}{\textbf{Overall}} \\
\cmidrule(lr){2-2}\cmidrule(lr){3-3}\cmidrule(l){4-6}
& \textbf{SR} $\uparrow$ & \textbf{SR} $\uparrow$ & \textbf{SR} $\uparrow$ & \textbf{SPL} $\uparrow$ & \textbf{Tokens/Ep.} $\downarrow$ \\
\midrule
Embedding & 0.776 & 0.282 & 0.224 & 0.079 & -- \\
VLN-Game (Dual) & 0.755 & 0.563 & 0.395 & 0.084 & 189.5 \\
\textbf{CoRelNav} & \textbf{0.816} & \textbf{0.631} & \textbf{0.513} & \textbf{0.129} & \textbf{91.2} \\
\bottomrule
\multicolumn{6}{@{}l}{\scriptsize Tokens are reported in thousands per episode.}
\end{tabular}
}
\end{table}

\begin{figure}[!t]
    \centering
    \makebox[\columnwidth][c]{%
        \includegraphics[width=\columnwidth]
        {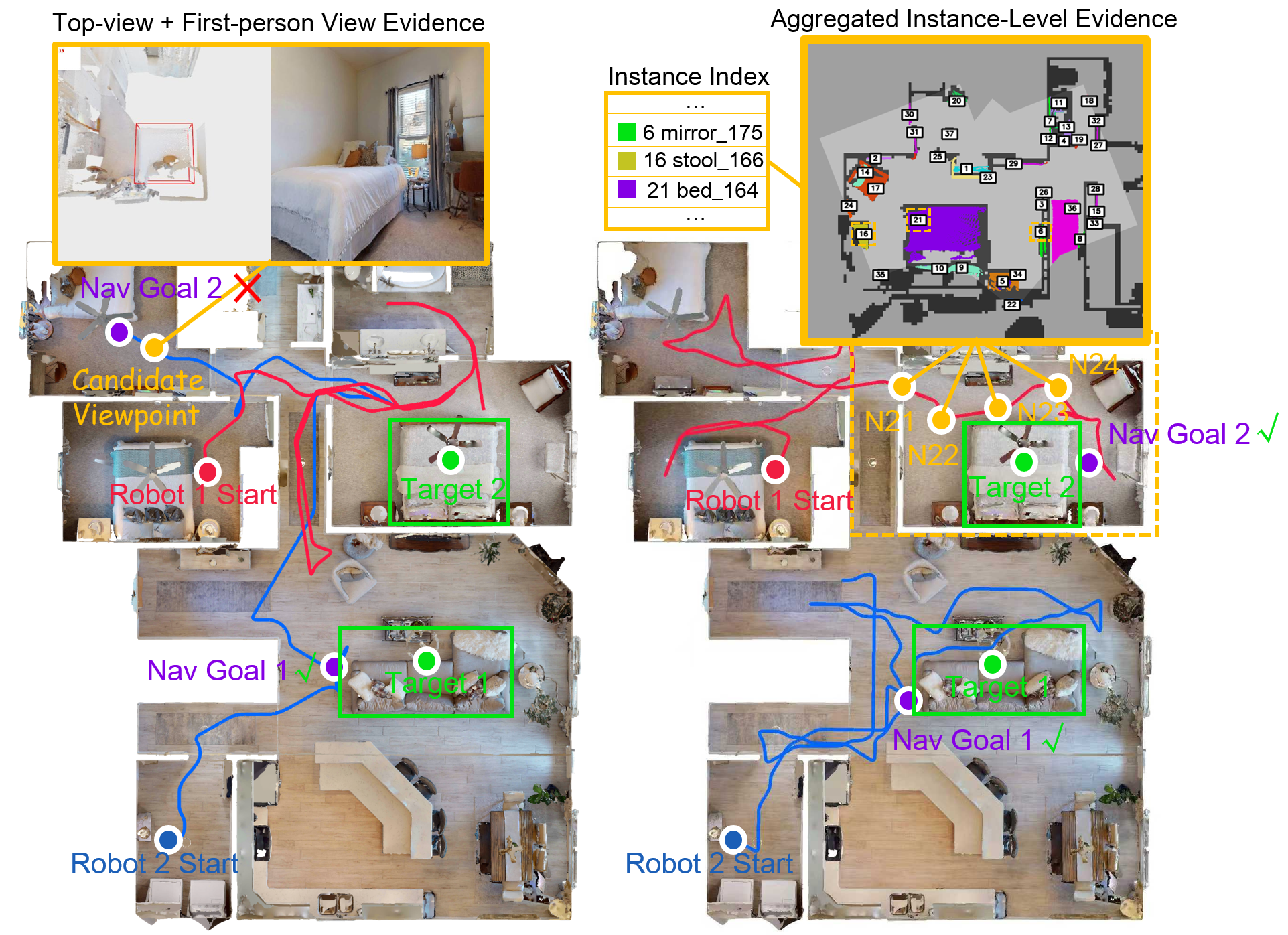}}

    \vspace{-1mm}
    \makebox[\columnwidth][c]{%
        \makebox[0.5\columnwidth][c]{\footnotesize (a) VLN-Game (Dual)}%
        \makebox[0.5\columnwidth][c]{\footnotesize (b) CoRelNav}}

    \caption{Comparison of VLN-Game (Dual) and CoRelNav on the same
    episode as Fig.~\ref{fig:simulation_qualitative}, with the instruction
    ``find a sofa and a bed between a stool and a mirror.'' Red and blue curves
    denote the trajectories of Robot 1 and Robot 2, respectively, and the
    labeled circles mark their start positions. Target 1 and Target 2 denote
    the ground-truth sofa and relational bed, while Nav Goal 1 and Nav Goal 2
    denote the final navigation goals selected by each method; $\checkmark$/$\times$
    indicate correct/incorrect target selection. (a) VLN-Game (Dual) verifies
    the relational candidate using a candidate-marked top view and first-person
    RGB evidence from the highlighted candidate viewpoint, but selects an
    incorrect bed. (b) CoRelNav aggregates instance-level evidence from topology
    nodes N21--N24, with object identities indexed in the inset, and
    correctly identifies the bed between the stool and the mirror.}
    \label{fig:dual_comparison}
\end{figure}

Table~\ref{tab:mixed_results} further highlights the distinction between
semantic discovery and relational grounding. The Embedding baseline performs
reasonably on object subtasks but degrades sharply on relational ones,
indicating that semantic similarity can identify plausible instances but
cannot determine whether they satisfy the required spatial constraints.
Explicit relational verification substantially reduces this gap.

CoRelNav achieves strong relational and episode-level performance by coupling task-conditioned exploration with persistent candidate hypotheses and structured multi-view evidence. Partial observations are retained and progressively completed, while candidate identities, local geometry, supporting objects, and task constraints are explicitly organized before relation reasoning and complementary observations of the same candidate are aggregated across topology nodes. This reduces the perceptual interpretation and instance association left to the language model, allowing it to focus on spatial-relation reasoning and reducing incorrect associations or hallucinated judgments. These mechanisms also focus language-model calls on task-relevant information, consistent with the lower token consumption in
Table~\ref{tab:mixed_results}. Compared with VLN-Game (Dual), which runs two
independent agents and verifies candidates mainly from a candidate-marked top
view and first-person RGB observations, CoRelNav additionally coordinates task
progress and evidence acquisition across robots and provides instance-numbered,
aggregated evidence for verification. As illustrated in
Fig.~\ref{fig:simulation_qualitative}, CoRelNav progressively evaluates multiple
bed candidates while completing the sofa subtask. Fig.~\ref{fig:dual_comparison}
further contrasts the resulting trajectories and verification evidence on the
same episode: VLN-Game (Dual) selects an incorrect relational navigation goal
from the highlighted candidate viewpoint, whereas CoRelNav aggregates
instance-consistent evidence from topology nodes N21--N24 and correctly reaches
the bed between the stool and the mirror.

\FloatBarrier

\subsubsection{Ablation Study}
We ablate three components: Single-Agent uses one robot with the same semantic
mapping and verification; w/o Joint Allocation removes joint frontier
assignment; and w/o Cross-node Evidence Aggregation verifies candidates from
individual topology nodes only.

\begin{table}[!ht]
\centering
\caption{Component ablations by relation complexity.}
\label{tab:ablation_results}
{\scriptsize\rmfamily
\setlength{\tabcolsep}{1.35pt}
\renewcommand{\arraystretch}{1.18}
\begin{tabular}{@{}cccccccc@{}}
\toprule
\multirow{2}{*}{\textbf{Variant}} & \multirow{2}{*}{\textbf{\#R}} & \multirow{2}{*}{\shortstack{\textbf{Joint}\\\textbf{Alloc.}}} & \multirow{2}{*}{\shortstack{\textbf{Cross-node}\\\textbf{Agg.}}} & \multicolumn{2}{c}{\textbf{1-Support}} & \multicolumn{2}{c}{\textbf{Multi-Support}} \\
\cmidrule(lr){5-6}\cmidrule(l){7-8}
& & & & \textbf{SR} $\uparrow$ & \textbf{SPL} $\uparrow$ & \textbf{SR} $\uparrow$ & \textbf{SPL} $\uparrow$ \\
\midrule
Single-Agent & 1 & -- & $\checkmark$ & 0.421 & 0.099 & 0.316 & 0.090 \\
w/o Joint Alloc. & 2 & $\times$ & $\checkmark$ & 0.526 & 0.116 & 0.395 & 0.102 \\
w/o Cross-node Agg. & 2 & $\checkmark$ & $\times$ & 0.553 & 0.141 & 0.342 & 0.105 \\
\textbf{CoRelNav} & 2 & $\checkmark$ & $\checkmark$ & \textbf{0.579} & \textbf{0.143} & \textbf{0.447} & \textbf{0.115} \\
\bottomrule
\end{tabular}
}
\end{table}

Table~\ref{tab:ablation_results} separates the roles of collaboration and
evidence integration. Joint Allocation consistently improves both relation
types, indicating that its primary contribution is to convert additional
robots into complementary rather than redundant exploration. Cross-node
Evidence Aggregation has a larger effect on multi-support relations, where
required evidence is more likely to span viewpoints; exact-instance
association allows these partial observations to remain bound to the same
hypothesis and be jointly verified. The two components are therefore
complementary: allocation acquires distributed evidence, while aggregation
makes it jointly usable.

On additionally constructed four-task mixed episodes, increasing the team
from two to three or four robots produced no further SPL gain, suggesting
diminishing path-efficiency returns from fleet size. Together with
Table~\ref{tab:ablation_results}, this indicates that CoRelNav benefits
primarily from task-aware coordination rather than increased team
size.

\FloatBarrier

\subsubsection{Limitations}
Three limitations remain. Candidate-specific evidence acquisition is activated only after a plausible target or supporting observation emerges, so missing or unobserved evidence may delay subsequent verification. Cross-node aggregation improves evidence completeness but cannot fully resolve ambiguous geometry or viewpoint-dependent errors in vision-language relation reasoning. Finally, simulation uses oracle masks and instance IDs to isolate exploration and reasoning; although the real-world system uses learned perception, robustness to detection, association, and calibration errors has not yet been systematically quantified.

\begin{figure}[!t]
    \centering

    \includegraphics[width=\columnwidth]
    {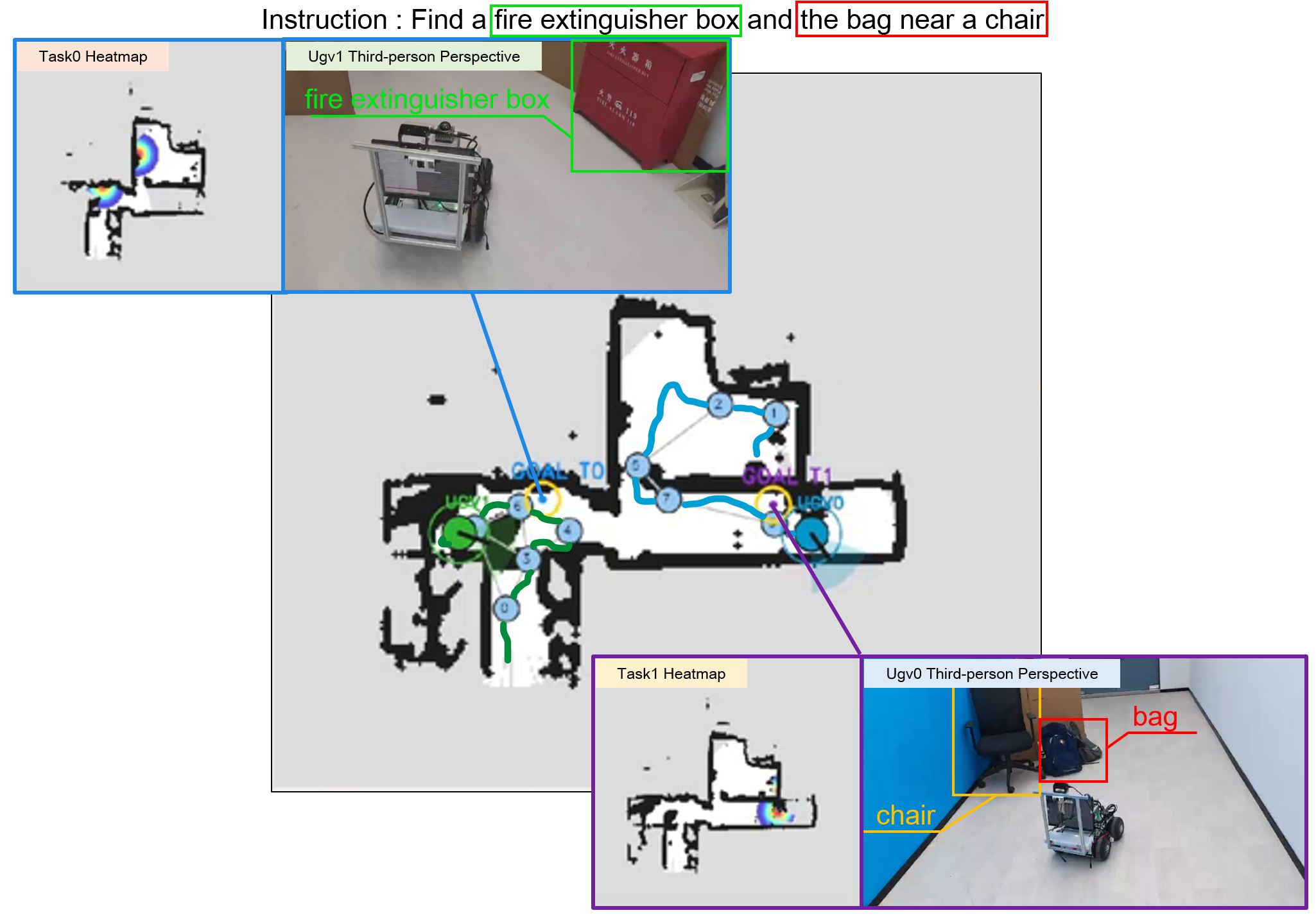}

    \caption{Representative real-world execution of
    ``find a fire extinguisher box and the bag near a chair.'' The shared map shows their coordinated trajectories, task goals, and topology nodes, while the insets present the corresponding task-conditioned heatmaps and third-person observations.}
    \label{fig:real_world_qualitative}
\end{figure}

\subsection{Real-World Experiment}
\label{sec:real_world_experiment}

We deploy CoRelNav on two wheeled robots in a previously unseen environment. Each robot carries an NVIDIA Jetson AGX Orin 64GB, a Livox MID-360 LiDAR, an Intel RealSense D435i RGB-D camera, and a SEEKER OMNI-4P panoramic camera. The onboard perception pipeline
uses Grounding DINO~\cite{liu2024groundingdino} for open-vocabulary object
detection and MobileSAM~\cite{zhang2023mobilesam} for instance masks. Camera–LiDAR extrinsic calibration associates detections with LiDAR measurements, forming semantic 3D point clouds registered by robot poses in a shared reference frame. The SEEKER OMNI-4P provides the panorama of each robot.

During deployment, the robots explore different regions, share maps and
semantic observations, and use the accumulated evidence to assign exploration
goals and verify candidate targets. Fig.~\ref{fig:real_world_qualitative}
presents a representative run in which the instruction requires locating a
fire extinguisher box and a bag near a chair. The two requirements are handled
as concurrent subtasks, and the resulting navigation waypoints are executed by
the local navigation systems. This example demonstrates the complete
perception, semantic mapping, collaborative exploration, and target
verification pipeline under real sensing and navigation conditions.

\section{Conclusion}
\label{sec:conclusion}

We presented CoRelNav, a collaborative framework for multi-robot navigation toward spatially constrained language goals in initially unknown environments. CoRelNav couples collaborative exploration and relational verification through task-conditioned spatial guidance and candidate-driven coordination. Task constraints, scene nodes, and object features guide exploration; candidate information, multi-view evidence, and team navigation costs adapt robot allocation and support relation verification. Simulations and ablations validate the design, and complete-system deployment on two physical robots demonstrates real-world feasibility. Future work will address larger teams, dynamic scenes, and robust spatial reasoning under imperfect perception.

\bibliographystyle{IEEEtran}
\bibliography{paper_refs}

@inproceedings{yamauchi1997frontier,
  title     = {A Frontier-Based Approach for Autonomous Exploration},
  author    = {Yamauchi, Brian},
  booktitle = {Proceedings of the IEEE International Symposium on Computational Intelligence in Robotics and Automation},
  pages     = {146--151},
  year      = {1997},
  doi       = {10.1109/CIRA.1997.613851}
}

@article{burgard2005coordinated,
  title   = {Coordinated Multi-Robot Exploration},
  author  = {Burgard, Wolfram and Moors, Mark and Stachniss, Cyrill and Schneider, Frank E.},
  journal = {IEEE Transactions on Robotics},
  volume  = {21},
  number  = {3},
  pages   = {376--386},
  year    = {2005},
  doi     = {10.1109/TRO.2004.839232}
}

@inproceedings{chaplot2020semexp,
  title     = {Object Goal Navigation Using Goal-Oriented Semantic Exploration},
  author    = {Chaplot, Devendra Singh and Gandhi, Dhiraj and Gupta, Abhinav and Salakhutdinov, Ruslan},
  booktitle = {Advances in Neural Information Processing Systems},
  volume    = {33},
  year      = {2020}
}

@inproceedings{ramakrishnan2022poni,
  title     = {{PONI}: Potential Functions for {ObjectGoal} Navigation with Interaction-Free Learning},
  author    = {Ramakrishnan, Santhosh Kumar and Chaplot, Devendra Singh and Al-Halah, Ziad and Malik, Jitendra and Grauman, Kristen},
  booktitle = {Proceedings of the IEEE/CVF Conference on Computer Vision and Pattern Recognition},
  pages     = {18890--18900},
  year      = {2022},
  doi       = {10.1109/CVPR52688.2022.01832}
}

@inproceedings{gadre2023cow,
  title     = {{CoWs} on {PASTURE}: Baselines and Benchmarks for Language-Driven Zero-Shot Object Navigation},
  author    = {Gadre, Samir Yitzhak and Wortsman, Mitchell and Ilharco, Gabriel and Schmidt, Ludwig and Song, Shuran},
  booktitle = {Proceedings of the IEEE/CVF Conference on Computer Vision and Pattern Recognition},
  pages     = {23171--23181},
  year      = {2023},
  doi       = {10.1109/CVPR52729.2023.02220}
}

@inproceedings{yokoyama2024vlfm,
  title     = {{VLFM}: Vision-Language Frontier Maps for Zero-Shot Semantic Navigation},
  author    = {Yokoyama, Naoki and Ha, Sehoon and Batra, Dhruv and Wang, Jiuguang and Bucher, Bernadette},
  booktitle = {Proceedings of the IEEE International Conference on Robotics and Automation},
  pages     = {42--48},
  year      = {2024},
  doi       = {10.1109/ICRA57147.2024.10610712}
}

@inproceedings{long2025instructnav,
  title     = {{InstructNav}: Zero-Shot System for Generic Instruction Navigation in Unexplored Environment},
  author    = {Long, Yuxing and Cai, Wenzhe and Wang, Hongcheng and Zhan, Guanqi and Dong, Hao},
  booktitle = {Proceedings of the 8th Conference on Robot Learning},
  series    = {Proceedings of Machine Learning Research},
  volume    = {270},
  pages     = {2049--2060},
  year      = {2025}
}

@inproceedings{yin2025unigoal,
  title     = {{UniGoal}: Towards Universal Zero-Shot Goal-Oriented Navigation},
  author    = {Yin, Hang and Xu, Xiuwei and Zhao, Linqing and Wang, Ziwei and Zhou, Jie and Lu, Jiwen},
  booktitle = {Proceedings of the IEEE/CVF Conference on Computer Vision and Pattern Recognition},
  pages     = {19057--19066},
  year      = {2025}
}

@article{yu2026vlngame,
  title   = {{VLN-Game}: Vision-Language Equilibrium Search for Zero-Shot Semantic Navigation},
  author  = {Yu, Bangguo and Liu, Yuzhen and Han, Lei and Kasaei, Hamidreza and Li, Tingguang and Cao, Ming},
  journal = {IEEE Transactions on Robotics},
  volume  = {42},
  pages   = {1824--1839},
  year    = {2026},
  doi     = {10.1109/TRO.2026.3677047}
}

@article{ortega2025divnav,
  title   = {{DIV-Nav}: Open-Vocabulary Spatial Relationships for Multi-Object Navigation},
  author  = {Ortega-Peimbert, Jes{\'u}s and Busch, Finn Lukas and Homberger, Timon and Yang, Quantao and Andersson, Olov},
  journal = {arXiv preprint arXiv:2510.16518},
  year    = {2025}
}

@inproceedings{jang2026contextnav,
  title     = {{Context-Nav}: Context-Driven Exploration and Viewpoint-Aware {3D} Spatial Reasoning for Instance Navigation},
  author    = {Jang, Won Shik and Kim, Ue-Hwan},
  booktitle = {Proceedings of the IEEE/CVF Conference on Computer Vision and Pattern Recognition},
  pages     = {9626--9636},
  year      = {2026}
}

@inproceedings{ye2022neuralcomapping,
  title     = {Multi-Robot Active Mapping via Neural Bipartite Graph Matching},
  author    = {Ye, Kai and Dong, Siyan and Fan, Qingnan and Wang, He and Yi, Li and Xia, Fei and Wang, Jue and Chen, Baoquan},
  booktitle = {Proceedings of the IEEE/CVF Conference on Computer Vision and Pattern Recognition},
  pages     = {14839--14848},
  year      = {2022}
}

@inproceedings{yu2022maans,
  title     = {Learning Efficient Multi-Agent Cooperative Visual Exploration},
  author    = {Yu, Chao and Yang, Xinyi and Gao, Jiaxuan and Yang, Huazhong and Wang, Yu and Wu, Yi},
  booktitle = {Proceedings of the European Conference on Computer Vision},
  pages     = {497--515},
  year      = {2022},
  doi       = {10.1007/978-3-031-19842-7_29}
}

@article{yu2025conavgpt,
  title   = {{Co-NavGPT}: Multirobot Cooperative Visual Semantic Navigation Using Vision Language Models},
  author  = {Yu, Bangguo and Yuan, Qihao and Li, Kailai and Kasaei, Hamidreza and Cao, Ming},
  journal = {IEEE Robotics and Automation Letters},
  volume  = {11},
  number  = {2},
  pages   = {2122--2129},
  year    = {2026},
  doi     = {10.1109/LRA.2025.3645650}
}

@inproceedings{shen2025mcoconav,
  title     = {Enhancing Multi-Robot Semantic Navigation Through Multimodal Chain-of-Thought Score Collaboration},
  author    = {Shen, Zhixuan and Luo, Haonan and Chen, Kexun and Lv, Fengmao and Li, Tianrui},
  booktitle = {Proceedings of the AAAI Conference on Artificial Intelligence},
  volume    = {39},
  number    = {14},
  pages     = {14664--14672},
  year      = {2025},
  doi       = {10.1609/aaai.v39i14.33607}
}

@article{james2026anygoal,
  title   = {{AnyGoal}: Vision-Language Guided Multi-Agent Exploration for Training-Free Lifelong Navigation},
  author  = {James, MoniJesu and Fernando, Marcelino Julio and Cabrera, Miguel Altamirano and Tsetserukou, Dzmitry},
  journal = {arXiv preprint arXiv:2606.13878},
  year    = {2026}
}

@inproceedings{huang2023vlmaps,
  title     = {Visual Language Maps for Robot Navigation},
  author    = {Huang, Chenguang and Mees, Oier and Zeng, Andy and Burgard, Wolfram},
  booktitle = {Proceedings of the IEEE International Conference on Robotics and Automation},
  pages     = {10608--10615},
  year      = {2023},
  doi       = {10.1109/ICRA48891.2023.10160969}
}

@inproceedings{gu2024conceptgraphs,
  title     = {{ConceptGraphs}: Open-Vocabulary {3D} Scene Graphs for Perception and Planning},
  author    = {Gu, Qiao and Kuwajerwala, Alihusein and Morin, Sacha and Jatavallabhula, Krishna Murthy and Sen, Bipasha and Agarwal, Aditya and Rivera, Corban and Paul, William and Ellis, Kirsty and Chellappa, Rama and Gan, Chuang and de Melo, Celso Miguel and Tenenbaum, Joshua B. and Torralba, Antonio and Shkurti, Florian and Paull, Liam},
  booktitle = {Proceedings of the IEEE International Conference on Robotics and Automation},
  pages     = {5021--5028},
  year      = {2024},
  doi       = {10.1109/ICRA57147.2024.10610243}
}

@inproceedings{chang2023ovsg,
  title     = {Context-Aware Entity Grounding with Open-Vocabulary 3D Scene Graphs},
  author    = {Chang, Haonan and Boyalakuntla, Kowndinya and Lu, Shiyang and Cai, Siwei and Jing, Eric Pu and Keskar, Shreesh and Geng, Shijie and Abbas, Adeeb and Zhou, Lifeng and Bekris, Kostas and Boularias, Abdeslam},
  booktitle = {Proceedings of the 7th Conference on Robot Learning},
  series    = {Proceedings of Machine Learning Research},
  volume    = {229},
  pages     = {1950--1974},
  year      = {2023}
}

@inproceedings{yang2025threedmem,
  title     = {{3D-Mem}: {3D} Scene Memory for Embodied Exploration and Reasoning},
  author    = {Yang, Yuncong and Yang, Han and Zhou, Jiachen and Chen, Peihao and Zhang, Hongxin and Du, Yilun and Gan, Chuang},
  booktitle = {Proceedings of the IEEE/CVF Conference on Computer Vision and Pattern Recognition},
  pages     = {17294--17303},
  year      = {2025},
  doi       = {10.1109/CVPR52734.2025.01612}
}

@inproceedings{savva2019habitat,
  title     = {Habitat: A Platform for Embodied {AI} Research},
  author    = {Savva, Manolis and Kadian, Abhishek and Maksymets, Oleksandr and Zhao, Yili and Wijmans, Erik and Jain, Bhavana and Straub, Julian and Liu, Jia and Koltun, Vladlen and Malik, Jitendra and Parikh, Devi and Batra, Dhruv},
  booktitle = {Proceedings of the IEEE/CVF International Conference on Computer Vision},
  pages     = {9339--9347},
  year      = {2019}
}

@article{ramakrishnan2021hm3d,
  title   = {{Habitat-Matterport 3D Dataset (HM3D)}: 1000 Large-scale {3D} Environments for Embodied {AI}},
  author  = {Ramakrishnan, Santhosh Kumar and Gokaslan, Aaron and Wijmans, Erik and Maksymets, Oleksandr and Clegg, Alexander and Turner, John and Undersander, Eric and Galuba, Wojciech and Westbury, Andrew and Chang, Angel X. and Savva, Manolis and Zhao, Yili and Batra, Dhruv},
  journal = {arXiv preprint arXiv:2109.08238},
  year    = {2021}
}

@article{anderson2018evaluation,
  title   = {On Evaluation of Embodied Navigation Agents},
  author  = {Anderson, Peter and Chang, Angel and Chaplot, Devendra Singh and Dosovitskiy, Alexey and Gupta, Saurabh and Koltun, Vladlen and Kosecka, Jana and Malik, Jitendra and Mottaghi, Roozbeh and Savva, Manolis and Zamir, Amir R.},
  journal = {arXiv preprint arXiv:1807.06757},
  year    = {2018}
}

@inproceedings{liu2024groundingdino,
  title     = {Grounding {DINO}: Marrying {DINO} with Grounded Pre-Training for Open-Set Object Detection},
  author    = {Liu, Shilong and Zeng, Zhaoyang and Ren, Tianhe and Li, Feng and Zhang, Hao and Yang, Jie and Jiang, Qing and Li, Chunyuan and Yang, Jianwei and Su, Hang and Zhu, Jun and Zhang, Lei},
  booktitle = {European Conference on Computer Vision},
  pages     = {38--55},
  year      = {2024},
  publisher = {Springer},
  doi       = {10.1007/978-3-031-72970-6_3}
}

@article{zhang2023mobilesam,
  title   = {Faster Segment Anything: Towards Lightweight {SAM} for Mobile Applications},
  author  = {Zhang, Chaoning and Han, Dongshen and Qiao, Yu and Kim, Jung Uk and Bae, Sung-Ho and Lee, Seungkyu and Hong, Choong Seon},
  journal = {arXiv preprint arXiv:2306.14289},
  year    = {2023}
}

@inproceedings{visser2013greedy,
  title     = {Discussion of Multi-Robot Exploration in Communication-Limited Environments},
  author    = {Visser, Arnoud and de Hoog, Julian and Jim{\'e}nez-Gonz{\'a}lez, Adrian and Mart{\'i}nez-de Dios, Jos{\'e} Ramiro},
  booktitle = {2013 ICRA Workshop on Towards Fully Decentralized Multi-Robot Systems: Hardware, Software and Integration},
  publisher = {Max Planck Institute for Biological Cybernetics},
  year      = {2013},
  url       = {https://dare.uva.nl/search?identifier=dff24be2-49af-4c6b-bdde-ecb4e8192478}
}

@article{julia2012costutility,
  title   = {A Comparison of Path Planning Strategies for Autonomous Exploration and Mapping of Unknown Environments},
  author  = {Juli{\'a}, Miguel and Gil, Arturo and Reinoso, {\'O}scar},
  journal = {Autonomous Robots},
  volume  = {33},
  number  = {4},
  pages   = {427--444},
  year    = {2012},
  doi     = {10.1007/s10514-012-9298-8}
}

@article{li2026qwen3vlembedding,
  title   = {{Qwen3-VL-Embedding} and {Qwen3-VL-Reranker}: A Unified Framework for State-of-the-Art Multimodal Retrieval and Ranking},
  author  = {Li, Mingxin and Zhang, Yanzhao and Long, Dingkun and Chen, Keqin and Song, Sibo and Bai, Shuai and Yang, Zhibo and Xie, Pengjun and Yang, An and Liu, Dayiheng and Zhou, Jingren and Lin, Junyang},
  journal = {arXiv preprint arXiv:2601.04720},
  year    = {2026}
}

@misc{alibaba2026qwen37plus,
  title        = {{Qwen3.7-Plus} Model Documentation},
  author       = {{Alibaba Cloud}},
  year         = {2026},
  howpublished = {Alibaba Cloud Model Studio documentation},
  url          = {https://help.aliyun.com/en/model-studio/qwen3-7-plus-us},
  note         = {Accessed: 2026-09-15}
}

\end{document}